# Estimation of Distribution Algorithms with Matrix Transpose in Bayesian Learning

Dae-Won Kim[1,a], Song Ko[1,b], and Bo-Yeong Kang[2,c,*]

[1]School of Computer Science and Engineering, Chung-Ang University, Seoul 156-756, Korea

[2]School of Mechanical Engineering, Kyungpook National University, Daegu 702-701, Korea

[a]dwkim@cau.ac.kr, [b]sko22.cau@gmail.com, [c]kby09@knu.ac.kr

**Keywords:** Estimation of distribution algorithms, Mutation, Bayesian network, Structure learning

**Abstract.** Estimation of distribution algorithms (EDAs) constitute a new branch of evolutionary optimization algorithms, providing effective and efficient optimization performance in a variety of research areas. Recent studies have proposed new EDAs that employ mutation operators in standard EDAs to increase the population diversity. We present a new mutation operator, a matrix transpose, specifically designed for Bayesian structure learning, and we evaluate its performance in Bayesian structure learning. The results indicate that EDAs with transpose mutation give markedly better performance than conventional EDAs.

**Introduction**

Estimation of distribution algorithms (EDAs) constitute a new branch of evolutionary optimization algorithms [1]; their workflow is similar to that of conventional GAs. After randomly reproducing chromosomes for the first generation, it repeats a set of genetic operations, i.e., selection, estimation, and reproduction, until a stopping criterion is fulfilled. The distinction between EDAs and GAs is based on the manner in which the genetic information is reproduced for offspring. In EDAs, a new population of individual solutions is generated by sampling a probabilistic model, which is estimated on the basis of representative individuals selected from the previous population. The advantages of EDAs over GAs are the absence of variation operators to be tuned and the expressiveness of the probabilistic model that drives the search process. Owing to these advantages, EDAs have been used as intuitive alternatives to GAs. Because the solutions of EDAs are evolved through a probabilistic model, the main issue is the construction of an effective probabilistic model. Many studies have proposed a variety of probabilistic models; they can be categorized into three approaches according to the manner of capturing the dependencies among variables: univariate (UMDA [2], PBIL [3], cGA [4]) bivariate (MIMIC [5], BMDA [6], DTEDA [7]), and multivariate (EcGA [8], EBNA [5], BOA [9]) approaches.

Bayesian networks are graphical structures for representing the probabilistic relationships among variables [10]. The structure learning of Bayesian networks is an NP-Hard optimization problem because the number of structures grows exponentially with the number of variables [11]. With regard to the application of EDAs for learning Bayesian networks, Blanco et al. used the UMDA and PBIL to infer the structure of Bayesian networks [12]. MIMIC was used to obtain the optimal ordering of variables for Bayesian networks [13]. The Bayesian networks learned using UMDA, PBIL, and MIMIC were more accurate than those learned using GAs, as compared to the original networks.

Recently, it has been reported that the incorporation of the mutation operator in EDAs can increase the diversity of genetic information in the generated population. However, the effectiveness of mutation-based EDAs in terms of the structure learning of Bayesian networks has not been investigated extensively. In this paper, we first present a new mutation operator, a matrix transpose, specifically designed for Bayesian structure learning. By exploiting the transpose mutation, we investigate the extent to which the performance of EDAs can be improved, and we try to determine the most improved EDA algorithm for learning Bayesian networks.

---

[*] Corresponding author

## PROPOSED METHOD

Recent studies have proposed new EDAs that employ mutation operators in standard EDAs to increase the population diversity. Handa used a bitwise mutation in UMDA, PBIL, MIMIC, and EBNA [14]; it was shown that the mutation operator improved the quality of solutions for the four-peaks problem, Fc4 function, and max-sat problem. Gosling et al. used a guided mutation in PBIL for the IPD strategy problem [15]; the mutation operator constrained the variation to solutions that were shown to be effective in the previous generation. Heien et al. compared the effectiveness of the bitwise mutation operator in BOA [16]; the mutation increased the success rate and reduced the minimum required population size in four function problems (Onemax, 5-trap, 3-deceptive, and 6-bipolar). Pelikan et al. analyzed the effects of bitwise mutation on improving the performance of UMDA through two test problems (Onemax and noisy Onemax) [17].

The benefits of mutation-based EDAs have inspired researchers to examine their effectiveness in Bayesian structure learning. Furthermore, conventional bitwise mutations are not closed operators from the viewpoint of the acyclicity of Bayesian networks; they can generate illegal solutions with cycles. In this study, we investigate the effectiveness of mutation in EDAs using a transpose mutation, designed for Bayesian structure learning; it enhances the diversity of the offspring and it increases the possibility of inferring the correct arc direction by considering the arc directions in candidate solutions as bi-directional, using the matrix transpose.

The choice of variation operators should follow intuitively from the problem representation. To represent a Bayesian network, we use a matrix representation, which is the most intuitive representation of a network. A network is represented as an $n \times n$ binary matrix M. The matrix element $M(i,j)$ in row $i$ and column $j$ is 1 if and only if variable $i$ is a parent of variable $j$ in the network. For example, the network in Fig. 1 (left) is represented in matrix form in Fig. 1 (right).

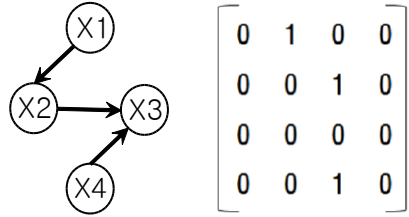

Fig. 1. Matrix representation of a Bayesian network.

Now, we build a probability matrix (P) to indicate the probability distribution of arcs among nodes in the selected individuals. Let $M = \{M_1, M_2, ..., M_d\}$ be a given population of individuals. Let $S = \{S_1, S_2, ..., S_h\}$ be a subset of individuals selected from M using their fitness ranks. Then, the $n \times n$ binary matrix P is defined by estimating the probability distribution of S. Using $h$ individuals, the occurrence frequency, and the average of each arc linking two nodes $i$ and $j$, the $P(i,j)$ is defined as $P(i,j) = 1/h \times (S_1(i,j) + S_2(i,j) + ... + S_h(i,j))$. $P(i,j)$ $(0 \leq P(i,j) \leq 1)$ represents the frequency with which an arc occurs in the selected individuals that were evaluated as promising individuals, and the importance of the arc in constructing an optimal structure.

For the population of next generation, the offspring O is generated by the probability matrix and transpose mutation until the number of offspring becomes $d$. To generate an element $O(i,j)$ of an offspring O, two probability values, i.e., $P(i,j)$ and a random number, are compared. Specifically, the element $O(i,j)$ is assigned a value of 1 if $P(i,j) > random[0,1]$; otherwise, it is assigned a value of 0.

In the case of a matrix representation, a variation operator has the choice of reversing the direction of arcs between two nodes with a matrix transpose; it generates offspring by inverting the arc direction in the individuals, which can inherit the information of solutions to drive the search over probable solutions and explore new states for offspring by changing the arc directions. Specifically, the transpose operator replaces $O(i,j)$ with $O(j,i)$ according to a mutation rate ($r$); $O(i,j)$ is assigned a value of $O(j,i)$ if $r > random[0,1]$; otherwise, it is assigned a value of $O(i,j)$. This matrix transpose mutation follows intuitively from the matrix representation, while preserving the necessary properties of the matrix, and it explores the conditional dependencies among variables in reverse order. A reversal operator has been widely used for the traveling salesman problem because of the possibility of avoiding local optima [18]. Moreover, the matrix transpose is a closed operator under Bayesian structure learning; illegal networks with cycles are not reproduced.

**RESULT**

We compare the performance of four standard EDAs (UMDA, PBIL, MIMIC, and BOA) with that of their mutation-adopted versions; the bitwise (EDAs+B) and transpose mutation (EDAs+T). The data sets employed were the Diabetes [19] and Asia [20] data; they have been widely used for comparative purposes in Bayesian structure learning. In these experiments, the well-known BDe score was used as the fitness function [10, 11].

The number of generations was set to 400; various mutation rates were used ($r = 0, 0.01, 0.05, 0.1, 0.15, 0.2$); five population sizes were used ($d = 10, 25, 50, 75, 100$); and five values of the learning parameter of the PBIL were used ($a = 0.1, 0.3, 0.5, 0.7, 0.9$). We designed $12 \times 5 \times 5 \times 5$ (12 EDAs, 5 populations sizes, 5 mutation rates, and 5 learning parameters) tests, and for each of these 1,500 configurations, we use 2 data sets, which gives us a total of 3,000 experiments. Each experiment was run 30 times.

Table 1. Comparison of precision (%) achieved by EDAs

| Data Set | Diabetes | | | | Asia | | | |
|---|---|---|---|---|---|---|---|---|
| Mutation | 0 | 0.01 | 0.05 | 0.10 | 0 | 0.01 | 0.05 | 0.10 |
| UMDA | 45.3 | - | - | - | 33.5 | - | - | - |
| UMDA+B | 45.3 | 46.4 | 47.3 | 54.3 | 33.5 | 32.1 | 32.6 | 51.3 |
| UMDA+T | 45.3 | 50.0 | 50.6 | 76.0 | 33.5 | 32.9 | 39.6 | 65.7 |
| PBIL | 42.0 | - | - | - | 35.3 | - | - | - |
| PBIL+B | 42.0 | 53.3 | 47.7 | 46.6 | 35.3 | 38.5 | 39.8 | 42.7 |
| PBIL+T | 42.0 | 63.8 | 78.9 | 79.9 | 35.3 | 45.3 | 60.8 | 70.1 |
| MIMIC | 41.8 | - | - | - | 27.8 | - | - | - |
| MIMIC+B | 41.8 | 45.9 | 46.9 | 44.6 | 27.8 | 26.9 | 30.4 | 37.2 |
| MIMIC+T | 41.8 | 43.3 | 44.5 | 47.9 | 27.8 | 27.8 | 28.5 | 38.5 |
| BOA | 41.2 | - | - | - | 27.7 | - | - | - |
| BOA+B | 41.2 | 42.0 | 42.3 | 34.4 | 27.7 | 27.1 | 24.4 | 30.6 |
| BOA+T | 41.2 | 39.3 | 48.0 | 56.1 | 27.7 | 26.4 | 30.5 | 44.1 |

Tables 1 lists the precisions achieved by each EDA for the two data sets for $r = 0, 0.01, 0.05$, and $0.10$ (with $d = 50$ and $a = 0.5$ fixed). The precision is the fraction of inferred arcs that are relevant to the network, which were assessed by comparing the network inferred by the EDAs with the original network. For the Diabetes data set, the bitwise mutation-adopted EDAs showed better performance than their standard versions; the EDAs+B with the best precision were 3~9% more accurate than their standard counterparts. The transpose mutation-adopted EDAs showed markedly better performance than their standard and bitwise versions, particularly those for UMDA+T and PBIL+T. In particular, PBIL+T ($r = 0.10$) achieved 79.9% precision; however, the results for PBIL and PBIL+B were precisions of 42.0% and 46.6% respectively.

For the Asia data set, EDAs+B provided higher precisions than their standard versions; UMDA+B showed the most improved performance. In contrast, the precisions of EDAs+T were superior to those of their standard and bitwise versions; the best precisions were approximately 10%~35% higher than those of the standard versions. UMDA+T and PBIL+T showed significantly better performance than UMDA/UMDA+B and PBIL/PBIL+B, giving best precisions of 65.7% and 70.1%, respectively. Of the transpose-adopted EDAs, MIMIC+T exhibited less improvement in performance.

Table 2. Proportion of arcs in the network inferred by EDAs

| Data Set | Arcs | EDAs | EDAs+B | EDAs+T |
|---|---|---|---|---|
| Diabetes | Correct | 42.6 | 45.0 | 65.0 |
| | Reverse | 43.9 | 41.3 | 26.4 |
| | Additional | 13.5 | 13.8 | 8.6 |
| Asia | Correct | 31.1 | 40.4 | 54.6 |
| | Reverse | 30.8 | 36.1 | 17.8 |
| | Additional | 38.1 | 23.4 | 27.6 |

Tables 2 compares the results of the average proportions of arcs in the network learned by EDAs, EDAs+B, and EDAs+T, for the Diabetes and Asia data, respectively. EDAs+T are superior to EDAs and EDAs+B; for inferring the correct arcs, EDAs+T were about 10%~23% more accurate than EDAs and EDAs+B. The present evaluation has verified the potential utility of the proposed method.